\documentclass[10pt,twocolumn,letterpaper]{article}

\usepackage[pagenumbers]{cvpr} 

\usepackage{graphicx}
\usepackage{amsmath}
\usepackage{amssymb}
\usepackage{booktabs}
\usepackage{url}

\usepackage{utfsym}
\usepackage{pifont}
\usepackage{soul} 
\usepackage{dsfont}

\usepackage{graphicx}
\usepackage{subcaption}
\usepackage{caption}

\usepackage[pagebackref,breaklinks,colorlinks]{hyperref}

\usepackage[capitalize]{cleveref}
\crefname{section}{Sec.}{Secs.}
\Crefname{section}{Section}{Sections}
\Crefname{table}{Table}{Tables}
\crefname{table}{Tab.}{Tabs.}

\makeatletter
\newcommand{\thickhline}{%
	\noalign {\ifnum 0=`}\fi \hrule height 1pt
	\futurelet \reserved@a \@xhline
}
\makeatother

\def\cvprPaperID{*****} 
\def\confName{CVPR}
\def\confYear{2025}

\begin{document}

\title{Leveraging Metric Depth for Relative Depth Prediction}  
\author{
Team BUPT MICLAB \\
Xiaoyang Bi$^{1,2}$,
Shuaikun Liu$^{1,2}$,
Zhaohong Liu$^{1,2}$,
Yuxin Yang$^{1,2}$,
Zhe Zhao$^{1,2}$,
Mengshi Qi$^{1,2}$\thanks{Corresponding author},\\
Liang Liu$^{1,2}$, and
Huadong Ma$^{1,2}$ \\
$^1$ Beijing Key Laboratory of Intelligent Telecommunications Software and Multimedia, \\
$^2$ Beijing University of Posts and Telecommunications \\
{\tt\small \{bxy, liushuaikun, liuzhaoh, yangyuxin, florasion, qms, liangliu, mhd\}@bupt.edu.cn}
}
\maketitle
\begin{abstract}
We present our solution to the 2025 SoccerNet Monocular Depth Estimation Competition Challenge. Predicting the relative depth in football scenarios is challenging, especially with only thousands of training samples available. To address this issue, our method leverages the powerful zero-shot capabilities of models pretrained on large-scale datasets to learn metric depth for effective relative depth prediction, achieving a score of $2.68 \times 10^{-3}$ on the challenge set.
\end{abstract}

\section{Introduction}
\label{sec:intro}
SoccerNet-Depth~\cite{leduc2024soccernet} is a recently introduced video-based monocular depth estimation (MDE) dataset covering both basketball and football scenes. Accurate depth estimation serves as a fundamental pillar for broader 3D scene understanding, with wide-ranging applications such as indoor depth completion~\cite{wang2022rgb,wang2024rdfc} and robust autonomous driving perception~\cite{liao2026improving,ye2025safedriverag,lv2025t2sg,zhu2023unsupervised}. In the specific context of sports videos, reliable spatial and depth cues are highly beneficial for advanced video understanding tasks, including 3D human pose estimation~\cite{qi2026towards}, action quality assessment~\cite{qi2025action,qi2025explainable}, temporal action localization~\cite{yun2024weakly}, and comprehensive multimodal video analysis (e.g., sports video captioning, retrieval, and classification)~\cite{qi2019sports,qi2021semantics,qi2020few}. The dataset provides relative depth annotations, as the depth data from the graphics engine encodes only relative, not absolute, values. In contrast, metric depth uses real-world units (e.g., meters), but for many vision tasks, the ordinal relationships between points (relative depth) are more important than absolute scale, allowing metric depth to serve as a proxy for relative depth in evaluation and training.

In this work, we finetune the Depth Anything model~\cite{yang2024depth}—a strong zero-shot MDE model pretrained on large-scale datasets—for metric depth estimation. Recently, large-scale foundational frameworks have demonstrated extraordinary zero-shot generalizability and robust feature representation across various physical reasoning and dense segmentation tasks~\cite{dc-sam,disentangled}. Motivated by these advancements, we treat the relative depth values in football games as metric depths within a 0–256 meter range. \textit{Since all samples are from football games with similar viewpoints}, the depth distribution is consistent, making it reasonable to interpret relative depths as metric values for finetuning and evaluation.

\section{Method}
In this section, we present the details of our method, including the network architecture, training objectives, and data augmentation strategies.

\subsection{Network Architecture}
\noindent\textbf{Depth Anything Encoder.}~The Depth Anything Model is a high-performance, zero-shot relative depth estimation model, which is built on the DinoV2~\cite{oquab2024dinov2} encoder and is pretrained on 62 million unlabeled images. It further enhances performance by incorporating semantic constraints from the DinoV2 model during training.

\noindent\textbf{ZoeDepth Decoder.}
Following~\cite{yang2024depth}, we freeze the powerful encoder of the Depth Anything Model and adopt a LocalBins-based decoder, improved by ZoeDepth~\cite{bhat2023zoedepth}, to convert relative depth predictions to metric depth. The decoder is randomly initialized and is the only trainable network module.
\subsection{Training objectives}
To supervise the model at the pixel level, we adopt the scale-invariant logarithmic loss~\cite{bhat2022localbins}, and omit chamfer loss for bin prediction due to its high memory cost and limited performance gain following ZoeDepth~\cite{bhat2023zoedepth}.

\subsection{Data augmentation}

We use only horizontal flipping for both train-time and test-time augmentation, as other methods like color jittering and random rotation reduced performance. At test time, we average predictions from the original and flipped images to improve results.


\section{Experiments}
In this section, we first present the implementation details, followed by an explanation of the rationale behind certain hyper-parameters and the data augmentation techniques used.
\subsection{Implementation Details}
The original image size of $1080 \times 1920$ is resized to $384 \times 704$ as the network input. The network outputs a prediction of the same size as the input, which is then upsampled to the original resolution using bilinear interpolation for evaluation. The entire network is optimized using the AdamW optimizer with a learning rate of $6.44 \times 10^{-4}$ and is trained for 100 epochs with a batch size of one on 8 RTX 4090 GPUs. The learning rate is kept constant for the first 70\% of epochs, after which it is reduced to $1/10,000$ of the initial value using cosine annealing.
\subsection{Results}
Our method achieves an RMSE of $2.68 \times 10^{-3}$ in the competition using DepthAnything as the backbone. During training, we observed that certain hyperparameters and data augmentation techniques have significant effects on performance.

\begin{figure}[t]
  \centering
   \includegraphics[width=0.8\linewidth]{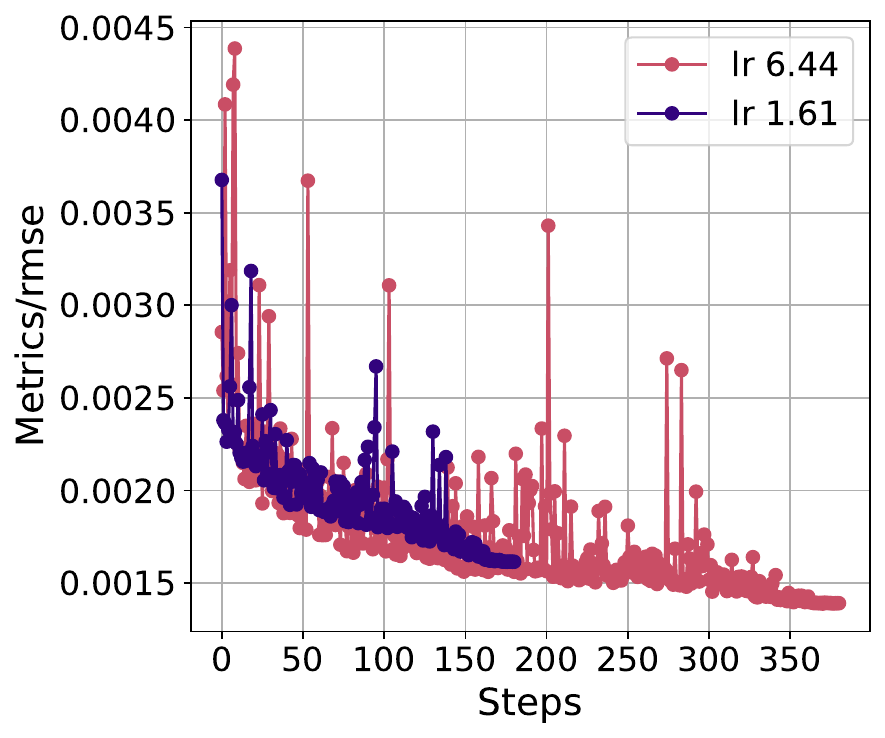}
   \caption{Visualization of the impact of learning rate and training epochs on the RMSE metric.} 
   \label{fig:lrcomp}
\end{figure}

\noindent\textbf{Learning Rate and Training Epoch.}~As shown in Table~\ref{table:comp}, we experimented with different learning rates and training epochs. The results indicate that a higher learning rate and longer training iterations are beneficial, reducing the RMSE from $1.60 \times 10^{-3}$ to $1.37 \times 10^{-3}$ on the 1/2 test set. Additionally, we visualize the evaluation results during training in Fig.~\ref{fig:lrcomp}, which shows that a lower learning rate results in fewer outliers but has less potential for further reduction. This observation inspired us to increase both the learning rate and the training schedule, leading to better convergence, as shown in the figure. Furthermore, we trained a DepthAnythingV2 model using the learning rate recommended by the authors, but it achieved worse results, suggesting that further hyperparameter tuning may be necessary.

\noindent\textbf{Train-time Data Augmentation.}~Table~\ref{table:abl} reports an ablation study on data augmentation, evaluated by RMSE on the full test set after 2 epochs. We restricted random rotation to $\pm 15^\circ$ and masked invalid regions after rotation when computing loss. However, rotation significantly worsened performance, and color jittering also had a slight negative effect, likely due to distribution mismatch with the test set. Thus, we only apply horizontal flipping for data augmentation during training.

\noindent\textbf{Test-time Data Augmentation.}~We also applied multi-scale test-time augmentation (with scales of 0.5, 1.0, and 1.5). As shown in row 7 of Table~\ref{table:abl}, this strategy led to a decrease in performance.

\noindent\textbf{Ensemble.}~Following~\cite{ke2024repurposing}, we performed an ensemble of two different models with similar RMSE performance. Specifically, we first aligned the scale and shift of the two predictions and then averaged their outputs. However, as shown in row 8 of Table~\ref{table:abl}, this method also resulted in degraded performance.

\textbf{
\begin{table}[h]
    \centering
    \vspace{-3mm}
    \small
    \begin{tabular}{cccc}
        \toprule
        Method & lr ($\times 10^{-4}$) & Training Epoch & RMSE ($\times 10^{-3}$) \\
        \midrule
        DA &  1.61 & 50 & 1.60 \\
        DA & 1.61  & 80 & 1.54 \\
        DA & 6.44 & 100 & \textbf{1.37} \\
        DAV2 & 0.4 & 100 & 2.63 \\
        \bottomrule
    \end{tabular}
    \caption{RMSE results obtained with different learning rates and training epochs. Evaluation is performed on half of the test set. DA(V2) denotes Depth Anything (V2) as the backbone.}
    \label{table:comp}
\end{table}
}

\textbf{
\begin{table}[h]
    \centering
    \vspace{-5mm}
    \small
    \begin{tabular}{cc}
        \toprule
        Method  &  RMSE ($\times 10^{-3}$) \\
        \midrule
        - & 4.09 \\
        +horizontal flipping & \textbf{3.50} \\
        +color jittering  & 4.04  \\
        +random rotation & 6.20  \\
        \midrule
        - & 1.39 \\
        +flipping test-time augmentation & \textbf{1.37} \\
        +multi-scale test-time augmentation  & 1.38  \\
        \midrule
        +ensemble & 1.42 \\
        \bottomrule
    \end{tabular}
\caption{Rows 1 to 4 show results for different train-time data augmentations (all models trained for 2 epochs). Row 5 to 7 show results for different test-time data augmentation. Row 8 shows the results with ensemble. The last four rows are evaluated on half of the test set.}
\label{table:abl}
\end{table}
}

\section{Conclusion}
In this report, we presented the method for learning from metric depth and directly predicting relative depth. We discussed various techniques, including hyperparameter tuning and augmentation strategies, and achieved an RMSE of $2.68 \times 10^{-3}$ in the 2025 SoccerNet Monocular Depth Estimation Competition Challenge.

{\small
\bibliographystyle{ieee_fullname}
\bibliography{egbib}
}

\end{document}